\def\year{2020}\relax
\documentclass[letterpaper]{article} 
\usepackage{aaai20}  
\usepackage{times}  
\usepackage{helvet} 
\usepackage{courier}  
\usepackage[hyphens]{url}  
\usepackage{graphicx} 
\usepackage{graphicx}  

\usepackage{amsmath}
\usepackage{amsfonts}
\usepackage{multirow}
\usepackage{bbding}
\usepackage{color}
\usepackage{nohyperref}

\title{Latent Opinions Transfer Network for Target-Oriented \\Opinion Words Extraction }
\author{Zhen Wu\protect\thanks{Authors contributed equally.}, Fei Zhao\footnotemark[1], Xin-Yu Dai\protect\thanks{Corresponding author.}, Shujian Huang, Jiajun Chen\\ 
	National Key Laboratory for Novel Software Technology, Nanjing University, Nanjing, 210023, China\\
	Collaborative Innovation Center of Novel Software Technology and Industrialization, Nanjing, 210023, China\\
	\{wuz, zhaof\}@smail.nju.edu.cn, \{daixinyu,huangsj,chenjj\}@nju.edu.cn 
}
 
\begin{document}

\maketitle

\begin{abstract}
	Target-oriented opinion words extraction (TOWE) is a new subtask of ABSA, which aims to extract the corresponding opinion words for a given opinion target in a sentence. Recently, neural network methods have been applied to this task and achieve promising results. However, the difficulty of annotation causes the datasets of TOWE to be insufficient, which heavily limits the performance of neural models. By contrast, abundant review sentiment classification data are easily available at online review sites. These reviews contain substantial latent opinions information and semantic patterns. In this paper, we propose a novel model to transfer these opinions knowledge from resource-rich review sentiment classification datasets to low-resource task TOWE. To address the challenges in the transfer process, we design an effective transformation method to obtain latent opinions, then integrate them into TOWE. Extensive experimental results show that our model achieves better performance compared to other state-of-the-art methods and significantly outperforms the base model without transferring opinions knowledge. Further analysis validates the effectiveness of our model.
\end{abstract}

\section{Introduction}
\begin{figure}[t]
	\centering
	\includegraphics[width=0.95\linewidth]{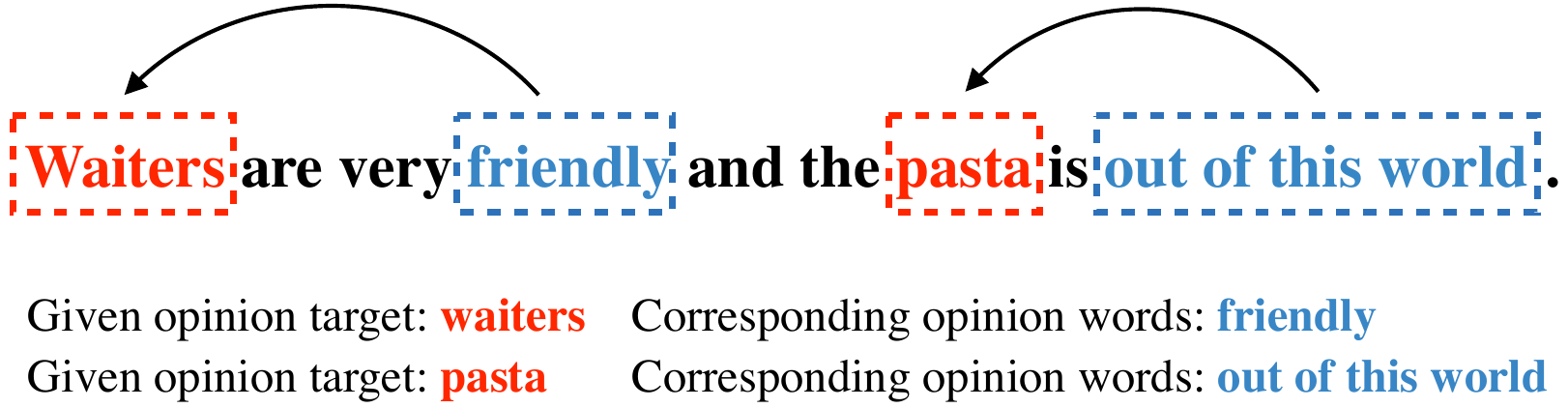}
	\caption{An example of TOWE. The words highlighted in red are two given opinion targets. TOWE task aims to extract the spans in blue as opinion words for the given targets. The arrows indicate the correspondence between opinion targets and opinion words. Note that opinion targets are given beforehand in the TOWE task.}
	\label{fig:towesample}
\end{figure}

Target-oriented opinion words extraction (TOWE)~\cite{DBLP:conf/naacl/FanWDHC19} is a new subtask of aspect-level sentiment analysis  (ABSA)~\cite{INR-011,DBLP:series/synthesis/2012Liu,DBLP:conf/semeval/PontikiGPPAM14}, which aims to extract the corresponding opinion words for a given opinion target from a review sentence. Opinion targets, also known as aspect terms, are the words or phrases in the sentence representing features or entities toward which users show attitudes. Opinion words refer to those terms of a sentence used to express attitudes or opinions explicitly. Figure~\ref{fig:towesample} shows an example of TOWE. In the sentence ``\emph{waiters are very friendly and the pasta is out of this world.}'', the terms ``\emph{waiters}'' and ``\emph{pasta}'' are two given opinion targets. TOWE needs to extract the word ``\emph{friendly}'' as the opinion word of the opinion target ``\emph{waiters}'' and the opinion words span ``\emph{out of this world}'' for the target ``\emph{pasta}''.

Many downstream sentiment analysis tasks, e.g., target-oriented sentiment classification~\cite{DBLP:conf/coling/TangQFL16,DBLP:conf/emnlp/WangHZZ16,DBLP:conf/acl/LiX18} and pair-wise opinion summarization~\cite{DBLP:conf/kdd/HuL04,DBLP:conf/cikm/ZhuangJZ06,DBLP:conf/coling/LiHHZXZY10}, can benefit from TOWE as it provides explicit opinion pairs information. To study this task,~\citeauthor{DBLP:conf/naacl/FanWDHC19}~\shortcite{DBLP:conf/naacl/FanWDHC19} released a benchmark corpus including four datasets and formalized TOWE as a problem of sequence labeling for given targets. Furthermore, they proposed a target-fused neural sequence labeling model and achieved state-of-the-art performance.

Despite the promising results of neural network methods, the lack of annotated data still heavily restricts the performance of TOWE. In practical scenarios, users usually refer to a considerable number of opinion targets in a review. It is extremely labor-intensive and time-consuming for annotators to identify all targets of a sentence and locate their corresponding opinion words. The difficulty of annotation causes the datasets of TOWE to be relatively scarce, which finally limits the effectiveness of neural models. In contrast, abundant labeled data of review sentiment classification are easily accessible at online review sites such as Amazon, Yelp and IMDB. Substantial opinions information and semantic patterns are naturally embodied in these reviews. Thus we propose to transfer them from large-scale sentiment classification datasets to the low-resource task TOWE. Although latent opinion knowledge is beneficial for TOWE, there are still two challenges remaining:

\begin{itemize}
	\item The opinions information such as opinion words in sentiment classification datasets are latent and unannotated, we need to find them explicitly before transferring them. 
	\item Since sentiment classification for reviews does not consider the target information, the latent-opinion information obtained is global and independent of the target. Thus, this information cannot be used directly by TOWE.
\end{itemize}

To address the above issues, we propose a novel model \textbf{L}atent \textbf{O}pinions \textbf{T}ransfer \textbf{N}etwork (LOTN) leveraging latent opinions knowledge from resource-rich review sentiment classification datasets to improve TOWE task. Specifically, we first pre-train an attention-based BiLSTM model on review sentiment classification datasets. The attention mechanism~\cite{DBLP:journals/corr/BahdanauCB14} is employed to extract possible opinion words through probabilistic weights. To solve the second issue, we design an effective transformation method to convert the global attention distribution over words in the sentiment classification model to latent target-dependent opinion words. Finally, we integrate these captured opinions words into TOWE model via an auxiliary learning signal. Additionally, we incorporate the encoder of the pretrained model to further guide TOWE model to learn latent opinions, which proves effective.

We evaluate the LOTN model on the four benchmark datasets. Results from extensive experiments indicate that our model achieves new state-of-the-art performance for TOWE and performs significantly better than our base model that does not transfer opinion knowledge. Further in-depth analysis also validates the effectiveness of our model. To the best of our knowledge, it is the first work to improve TOWE by transferring latent opinions knowledge of review sentiment classification datasets.

The main contributions of this work include:
\begin{itemize}
	\item In tackling the problem of insufficient annotated data, we are the first to propose transferring latent opinion knowledge from resource-rich review sentiment classification datasets to the low-resource task of TOWE.
	\item To transfer opinion information effectively, we propose a novel model that obtains latent opinion words from a sentiment classification model and integrates them into TOWE via an auxiliary learning signal.
	\item The experiment results indicate that our model achieves better results compared to state-of-the-art methods. Extensive analysis validates the effectiveness of our model.
\end{itemize}

\section{Preliminary}
In this section, we will introduce the task formalization of TOWE and the pretrained sentiment classification model that is used for transferring latent opinions.
\subsection{TOWE Formalization}
\begin{table}[ht]
	\small
	\centering
	\caption{Different labeling results of a sentence when given different opinion targets. The opinion targets are highlighted in underline and the opinion words/phrases are in bold.}
	\label{taskdefinition}
	\begin{tabular}{c p{7.2cm}}
		\hline
		1. & \textrm{\underline{Waiters}/O are/O very/O \textbf{friendly}/B and/O the/O pasta/O is/O out/O of/O this/O world/O ./O} \\
		2. & \textrm{Waiters/O are/O very/O friendly/O and/O the/O \underline{pasta}/O is/O \textbf{out}/B \textbf{of}/I \textbf{this}/I \textbf{world}/I ./O} \\
		\hline
	\end{tabular}
\end{table}
TOWE aims to extract the corresponding opinion words for a given target from a sentence, which can be formalized as a task of sequence labeling for given targets. Specifically, given a review sentence $s=\{w_1, w_2, \cdots, w_n\}$ consisting $n$ words and an opinion target $w_t$  in the sentence $s$ (\textbf{Note that}, we notate an opinion target as one word for simplicity and $t$ is the position of the target in the sentence), the goal is to tag each word $w_i$ in $s$ with a label $y_i \in \{B, I, O \}$ (B: Beginning, I: Inside, O: Others). The spans composed by the tags $B$ and $I$ represent the corresponding opinion words of the target $w_t$. For example, the sentence in Figure~\ref{fig:towesample} is tagged as $w_i/y_i$ for different opinion targets as shown in Table~\ref{taskdefinition}.
\begin{figure*}[t]
	\centering
	\includegraphics[width=0.8\textwidth]{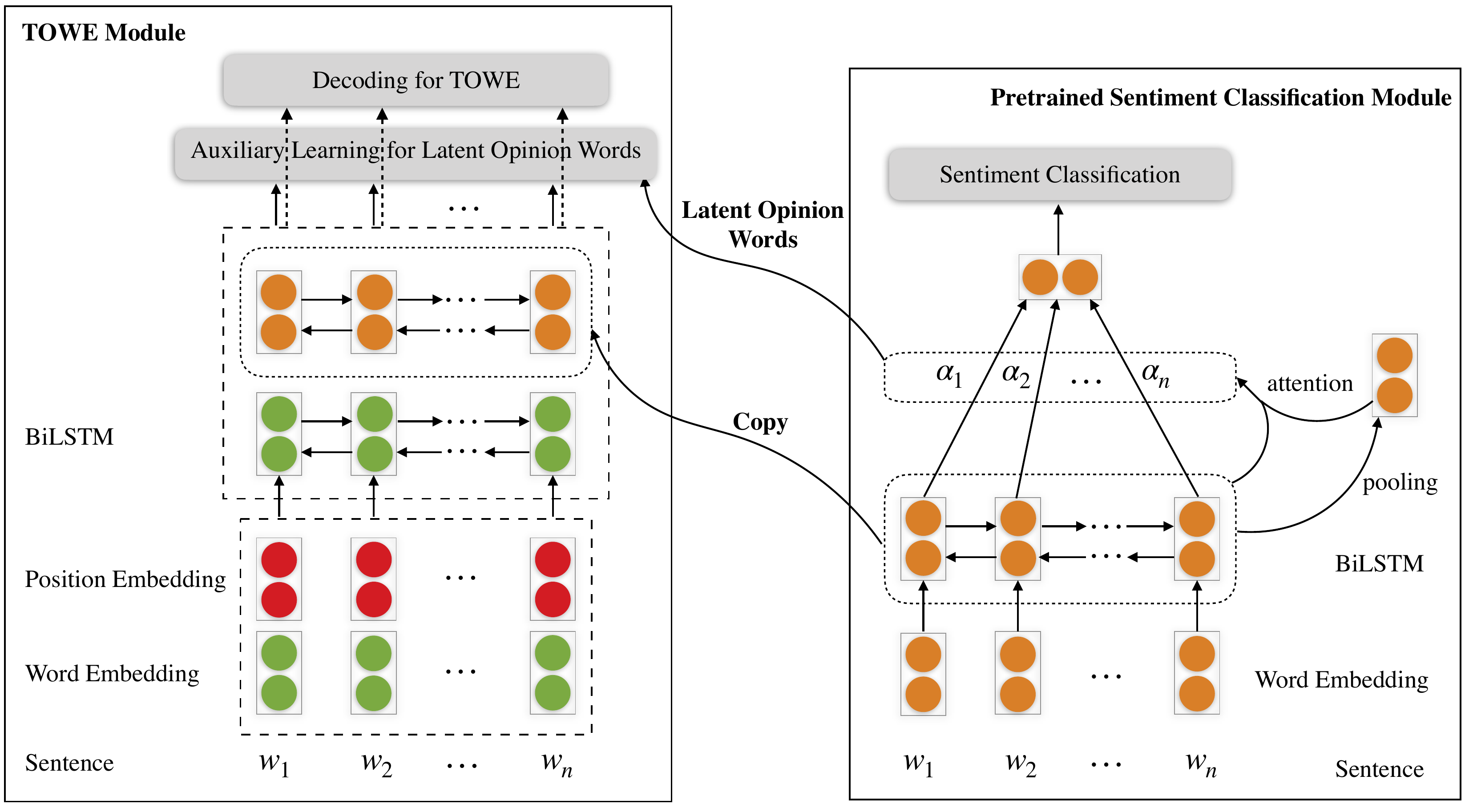}
	\caption{The architecture of Latent Opinions Transfer Network. Different opinion targets in a sentence have different position embeddings.}
	\label{fig:model}
\end{figure*}

\subsection{Pretraining Sentiment Classification Model}
Review sentiment classification aims to detect overall sentiment polarity (e.g., positive or negative) of a review text. Before transferring latent opinions, we first pretrain an attention-based BiLSTM model on large-scale review sentiment classification datasets.

Specifically, we regard a review from sentiment classification datasets as a long sentence $\{w_1, w_2, \cdots, w_m\}$ consisiting $m$ words, and map them into the corresponding vector representations $\{\mathbf{w}_1, \mathbf{w}_2, \cdots, \mathbf{w}_m\}$ by looking up an embedding table. Then a BiLSTM network is applied to encode the word representations $\{\mathbf{w}_1, \mathbf{w}_2, \cdots, \mathbf{w}_m\}$ and genenrate the context representations $\{\mathbf{h}_1^{sc}, \mathbf{h}_2^{sc}, \cdots, \mathbf{h}_m^{sc}\}$.

The attention mechanism is employed to capture the latent and global opinion words that are significant to sentiment classification. The attention weight $\alpha_i$ of $\mathbf{h}_i^{sc}$ is defined as:
\begin{align}
	u(\mathbf{h}_i^{sc}, \mathbf{h}_{avg}^{sc} ) &= \mathbf{h}_i^{sc} \cdot \mathbf{W}_{u} \cdot \mathbf{h}_{avg}^{sc} + b_{u}, \\
	\alpha_i &= \frac{\exp(u(\mathbf{h}_i^{sc}, \mathbf{h}_{avg}^{sc} ))}{\sum_{j=1}^{m} \exp(u(\mathbf{h}_j^{sc}, \mathbf{h}_{avg}^{sc}) )},
\end{align}
where $\mathbf{h}_{avg}^{sc}$ is the average of all hidden states, i.e., $\mathbf{h}_{avg}^{sc} = \sum_{j=1}^{m} \mathbf{h}_{j}^{sc}/m$, $\mathbf{W}_{u}$ and $b_u$ are the weight matrix and bias.

The review representation $\mathbf{r}_{sc}$ is a weighted sum of all hidden states:
\begin{equation}
	\mathbf{r}_{sc} = \sum_{i=1}^{m}\alpha_i \mathbf{h}_i^{sc}.
\end{equation}

Finally, the representation $\mathbf{r}_{sc}$ is fed to a linear layer and a softmax layer to predict the sentiment label of the review. We train the sentiment classification model by minimizing the cross-entropy loss between the predicted sentiment distribution and the ground truth. After pretraining is finished, all parameters in sentiment classification model are fixed.

\section{Latent Opinions Transfer Network}
In this section, we introduce our model Latent Opinions Transfer Network in details. We will present the overall architecture of the model, our base TOWE model and two proposed latent opinions transfer methods, as well as the final decoding and training.

\subsection{Overall Description}
Figure~\ref{fig:model} shows the overall architecture of Latent Opinions Transfer Network (LOTN). It consists of two components mainly: a TOWE module and a pretrained review sentiment classification module. We design a simple and effective network as our base TOWE module,  namely position embedding based BiLSTM (PE-BiLSTM). The pretrained sentiment classification module is the aforementioned attention-based BiLSTM network. LOTN transfers latent opinions from the sentiment classification module to the TOWE module through two different perspectives.

Firstly, the BiLSTM layer of the pretrained sentiment classification module contains substantial implicit opinion information and semantic patterns. We integrate this information into the encoding layer of TOWE module to introduce external opinion knowledge.

Secondly, the latent opinion words captured by the attention process of the pretrained module are global and target-independent since review sentiment classification does not consider the target information. To address this issue, we propose a heuristic transformation method to convert global attention weights over words to latent target-dependent opinion words by considering the position information of the target and other words. Then we incorporate them into the TOWE module via an auxiliary learning signal.

\subsection{Position Embedding based BiLSTM}
In the base model, we use position embedding to model the target information instead the state-of-the-art model IOG (Fan et al. 2019) because IOG employs six different positional and directional LSTMs simultaneously and suffers from high model complexity. In contrast to IOG, position embedding is a simple and effective method of modeling position information and widely used in Natural Language Processing~\cite{DBLP:conf/acl/LinSLLS16,DBLP:conf/acl/GehringAGD17,DBLP:conf/nips/VaswaniSPUJGKP17,DBLP:conf/coling/GuZHS18}.

Given the sentence  $s=\{w_1, w_2, \cdots, w_n\}$ and the opinion target $w_t$ in the sentence, we first generate the relative distance index $l_i=|i-t|$ of each word $w_i$ in $s$ by calculating the relative distance from $w_i$ to the target $w_t$. Then the distance index $l_i$ is mapped into the positional representation by using a position embedding tabel  $\mathbf{E}_{pos}\in \mathbb{R}^{L\times d_1}$, where $d_1$ is the embedding dimension and $L$ is the maximal position index. In addition, we also use a word embedding table $\mathbf{E}_{emb}\in \mathbb{R}^{|V|\times d_2}$ to obtain the semantic representation of word. The representation $\mathbf{e}_i$ of each word $w_i$ is formed by concatenating the word vector and the corresponding position vector:
\begin{equation}
	\mathbf{e}_i = \left [\mathbf{E}_{emb}(w_i); \mathbf{E}_{pos}(l_i) \right ],
\end{equation}
where the $\left [\cdot;\cdot \right ]$ denotes the vector concatenation operation.

Finally, we employ a BiLSTM network to capture the contextual information of each word. The simplified update rule can be written as follows:
\begin{equation}
	\mathbf{h}_i^t = \text{BiLSTM}(\mathbf{h}_{i-1}^t, \mathbf{e}_i, \theta_t ),
\end{equation}
where $\theta_t$ represents the parameters of BiLSTM.

In the base model, the context representation $\mathbf{h}^t$ can be used for predicting the opinion words of the given target.

\subsection{Transferring Pretrained Encoder}
In order to transfer latent opinions knowledge, we also feed the sentence $s$ of TOWE task to the pretrained sentiment classification module to yield the corresponding hidden states $\{\mathbf{h}^{sc}_1, \mathbf{h}^{sc}_2, \cdots, \mathbf{h}^{sc}_n\}$ and attention weights $\{\alpha_1, \alpha_2, \cdots, \alpha_n\}$.

From the semantic level view, the encoder of the pretrained sentiment classification module holds substantial implicit opinion information, and thus we integrate it into the TOWE module by concatenating two hidden states:

\begin{equation}
	\mathbf{r}_i = \left [ \mathbf{h}^t_i; \mathbf{h}^{sc}_i \right ],
\end{equation}
where $\mathbf{r}_i$ contains both task-specific context representations and external opinions knowledge from review sentiment classification datasets.

\subsection{Transferring  Latent Opinion Words}
To effectively transfer latent opinion words from sentiment classification module to TOWE module, we design a heuristic transformation method and an auxiliary learning signal respectively to capture and integrate them into TOWE.
\subsubsection{Transformation Method}
As we have mentioned, the attention mechanism in the sentiment classification module captures latent opinion words in the style of probabilistic weights. However. these probabilistic opinions information are global and target-independent. Intuitively, the word that is closer to the opinion target is more likely to be the opinion word of the target. Thus we introduce the opinion target information into the attention distribution by a target-relevant distance weight $c_i$:
\begin{align}
	\alpha_i{'} &= c_i \cdot \alpha_i, \\
	c_i &= 1 - \frac{|i-t|}{n},
\end{align}
where $n$ is the length of input sentence, $t$ indicates the position of the opinion target $w_t$ in the sentence, and $|i-t|$ denotes the relative distance between the word $w_i$ and the target $w_t$. It can be observed that a closer word to the target has a bigger distance weight. To regain the probabilistic attention distribution, the target-dependent attention weight $\alpha_i{'}$ is re-normalized:
\begin{equation}
	\beta_i=\frac{\alpha_i{'}}{\sum_{j=1}^{n}\alpha_j{'}}.
\end{equation}

Finally, we use a heuristic strategy to convert the normalized attention weight $\beta_i$ into the binary latent opinion words by the threshold $\frac{1}{n}$:
\begin{equation}
	y^a_i=\begin{cases}
		1 & \text{ if } \beta_i \geq  \frac{1}{n}, \\ 
		0 & \text{ otherwise }, 
	\end{cases}
\end{equation}
where $y^a_i=1$ denotes that the word $w_i$ is a latent and target-relevant opinion word from the perspective of sentiment classification module and $0$ indicates not.

\subsubsection{Auxiliary Learning Signal}
In fact, the $y^a_i$ is a pseudo label and represents the opinions knowledge from the sentiment classifcation module. We integrate these latent opinions into TOWE module by auxiliary learning signal:
\begin{align}
	\hat{\mathbf{y}}^a_i&=\rm{softmax}(\mathbf{W}_a\mathbf{r}_i+\mathbf{b}_a),\\
	\mathcal{L}_a &=-\sum_{i=1}^{n}\sum_{k=0}^{1}\mathbb{I}(y^a_i=k)\log(\hat{y}^a_{i,k}),
\end{align}
where $\mathbf{W}_a$ and $\mathbf{b}_a$ are the weight matrix and  the bias, $\hat{\mathbf{y}}^a_i$ represents the prediction probability and $\mathbb{I}(\cdot)$ is the indicator function. LOTN embraces these latent opinions knowledge by optimizing the auxiliary loss $\mathcal{L}_a$, which helps TOWE moduel decode the opinion words of the target better.

\subsection{Decoding and Training}
Since the context representation $\mathbf{r}_i$ contains both task-specfic opinions and tranferred opinions knowledge, LTON finally use $\mathbf{r}_i$ to predict the tag $y_i\in\{B,I,O\}$ of the word $w_i$. It can regarded as a three-class classfication problem at each position of the sentence $s$. We use a linear layer and a softmax layer to compute prediction probaility $\hat{\mathbf{y}}_i$:
\begin{equation}
	\hat{\mathbf{y}}_i=\rm{softmax}(\mathbf{W}_t\mathbf{r}_i+\mathbf{b}_t),
\end{equation}
where $\mathbf{W}_t$ is the weight matrix and $\mathbf{b}_t$ denotes the bias. The cross-entropy loss of TOWE task can be defined as follows:
\begin{equation}
	\mathcal{L}_t=- \sum_{i=1}^{n}\sum_{k=0}^{2}\mathbb{I}(y_i=k)\log(\hat{y}_{i,k}),
\end{equation}
here the tags $\{O, B, I\}$ are correspondingly converted into labels $\{0, 1, 2\}$ and $y_i$ denotes the ground truth label.

LOTN also integrates latent opinions through auxiliary learning signal $\mathcal{L}_a$. Thus the final loss is defined as follows:
\begin{equation}
	J=\mathcal{L}_t+\lambda \mathcal{L}_a,
	\label{finalloss}
\end{equation}
where $\lambda$ measures the importance of auxiliary learning and can be adjusted. We minimize the loss $J$ to optimize the LOTN model.

\section{Experiments}
\subsection{Datasets and Metrics}
We evaluate our model on four benchmark datasets~\cite{DBLP:conf/naacl/FanWDHC19}. The statistics of the datasets are summarized in Table~\ref{tab:towedatasets}. The datasets 14res and 14lap are derived from SemEval Challenge 2014 task 4~\cite{DBLP:conf/semeval/PontikiGPPAM14}, 15res and 16res are respectively from SemEval Challenge 2015 task 12~\cite{DBLP:conf/semeval/PontikiGPMA15} and SemEval Challenge 2016 task 5~\cite{DBLP:conf/semeval/PontikiGPAMAAZQ16}. The suffixes ``res'' and ``lap'' respectively represent reviews from restaurant domain or laptop domain. The original SemEval Challenge datasets are very popular benchmarks for ABSA subtasks. They provide the annotation of opinion targets for each sentence, but not the corresponding opinion words. To perform TOWE task,~\citeauthor{DBLP:conf/naacl/FanWDHC19}~\shortcite{DBLP:conf/naacl/FanWDHC19} annotate the corresponding opinion words for the given targets from sentences and remove the cases without explicit opinion words.

\begin{table}[!htbp]
	\small
	\centering
	\caption{Statistics of TOWE datasets. A sentence may contain multiple opinion targets.}
	\label{tab:towedatasets}
	\begin{tabular}{cc|cc}
		\hline
		\multicolumn{2}{c|}{Datasets} & {\#sentences} & {\#targets}\\
		\hline
		\multirow{2}*{14res}  & Train & 1,627 & 2,643 \\
		& Test & 500 & 864\\
		\hline
		\multirow{2}*{14lap}  & Train & 1,158 & 1,634\\
		& Test & 343 & 482 \\
		\hline
		\multirow{2}*{15res}  & Train & 754 & 1,076\\
		& Test & 325 & 436 \\
		\hline
		\multirow{2}*{16res}  & Train & 1,079 & 1,512\\
		& Test & 329 & 457 \\
		\hline		
		
	\end{tabular}
\end{table}

To pretrain the sentiment classification model, we use the two datasets respectively from Amazon Review and Yelp Review. The Yelp Review is applied to transfer latent opinions for TOWE datasets 14res, 15res, and 16res. The Amazon Review is used for the dataset 14lap. Table~\ref{tab:reviewdatasets} shows the statistics of Amazon Review and Yelp Review.

\begin{table}[!htp]
	\small
	\centering
	\caption{Statistics of the two datasets Amazon Review and Yelp Review.}
	\label{tab:reviewdatasets}
	\begin{tabular}{c|cccc}
		\hline
		{Datasets} & {\#positive} & {\#negative} & {\#total} \\
		\hline
		Yelp Review & 266,041 & 177,218 & 443,259 \\
		Amazon Review & 277,228 & 277,769 & 554,997 \\
		\hline
	\end{tabular}
\end{table}

Following the state-of-the-art work~\cite{DBLP:conf/naacl/FanWDHC19}, we use the metrics precision, recall, and F1-score to measure the performance of different methods. An opinion word/phrase is deemed to be correct on the condition that the starting and ending positions of the prediction are both the same as those of the golden word/phrase.

\subsection{Experiment Settings}
We initialize word vectors with 300-dimension word embeddings from GloVe~\cite{DBLP:conf/emnlp/PenningtonSM14}. The word vectors are fixed and not fine-tuned during the training stage. We set the dimension of position embeddings to be 300. The position embeddings, all weight matrices and biases are randomly initialized by a uniform distribution $U(-0.01, 0.01)$. The dimension of LSTM cell is set to 200. We adopt Adam optimizer~\cite{DBLP:journals/corr/KingmaB14} to update parameters of models. The initial learning rate is 0.001 and mini-batch size is set to 25. The dropout ~\cite{DBLP:journals/corr/abs-1207-0580} is applied on embedding layer with probability 0.5. We randomly select 20\% of training set as validation set for tuning hyper-parameters and early stopping. Finally, we run each model five times and report the average result of them.

\subsection{Compared Methods}
\begin{table*}[htbp]
	\caption{Main experiment results(\%). Best results are in bold (P, R, and F1-score, the larger is the better). The marker${\  }^{\dagger}$ represents that LOTN outperforms other methods significantly ($p < 0.01$) .} 
	\label{mainresults}
	\resizebox{0.98\textwidth}{!}{
		\begin{tabular}{c|c c c|c c c|c c c|c c c}
			\hline
			\multirow{2}{*}{Models} & \multicolumn{3}{c|}{14res} & \multicolumn{3}{c|}{14lap} & \multicolumn{3}{c|}{15res} & \multicolumn{3}{c}{16res}\\ 
			\cline{2-13}
			&P& R& F1 &P& R& F1&P& R& F1&P& R& F1 \\
			\hline
			Distance-rule & 58.39&43.59&49.92 & 50.13&33.86&40.42 & 54.12&39.96&45.97 & 61.90&44.57&51.83\\ 
			
			Dependency-rule & 64.57&52.72&58.04 & 45.09&31.57&37.14 & 65.49&48.88&55.98 & 76.03&56.19&64.62\\
			
			LSTM & 52.64&65.47&58.34 & 55.71&57.53&56.52 & 57.27&60.69&58.93 & 62.46&68.72&65.33\\ 
			
			BiLSTM & 58.34&61.73&59.95 & 64.52&61.45&62.71 & 60.46&63.65&62.00 & 68.68&70.51&69.57\\
			
			Pipeline & 77.72&62.33&69.18 & 72.58&56.97&63.83 & 74.75&60.65&66.97 & 81.46&67.81&74.01\\
			
			TC-BiLSTM & 67.65&67.67&67.61 & 62.45&60.14&61.21 & 66.06&60.16&62.94 & 73.46&72.88&73.10\\ 
			
			IOG & 82.38& 78.25& 80.23 & 73.43& \textbf{68.74}& 70.99 & 72.19 &  \textbf{71.76} & 71.91 & 84.36 & 79.08 & 81.60 \\
			
			\hline
			PE-BiLSTM & 80.10 & 76.51 & 78.26 & 72.01&64.20& 67.83 & 70.36 & 65.73 & 67.96 & 82.27 & 74.95 & 78.43 \\
			LOTN & $\textbf{84.00}^{\dagger}$ & $\textbf{80.52}^{\dagger}$ &  $\textbf{82.21}^{\dagger}$ &  $\textbf{77.08}^{\dagger}$ & 67.62&  $\textbf{72.02}^{\dagger}$ &  $\textbf{76.61}^{\dagger}$ &  70.29 &  $\textbf{73.29}^{\dagger}$ &  $\textbf{86.57}^{\dagger}$ &  $\textbf{80.89}^{\dagger}$ &  $\textbf{83.62}^{\dagger}$ \\
			
			\hline
	\end{tabular} }
	\centering
\end{table*}

\begin{table*}[htbp]
	\caption{Experiment results of adding the transferred encoder or auxiliary learning on PE-BiLSTM(\%).} 
	\label{ablationstudy}
	\resizebox{0.98\textwidth}{!}
	{
		\begin{tabular}{c|c c c|c c c|c c c|c c c}
			\hline
			\multirow{2}{*}{Models} & \multicolumn{3}{c|}{14res} & \multicolumn{3}{c|}{14lap} & \multicolumn{3}{c|}{15res} & \multicolumn{3}{c}{16res}\\ 
			\cline{2-13}
			&P& R& F1 &P& R& F1&P& R& F1&P& R& F1 \\
			\hline
			PE-BiLSTM & 80.10 & 76.51 & 78.26 & 72.01&64.20& 67.83 & 70.36 & 65.73 & 67.96 & 82.27 & 74.95 & 78.43 \\
			\hline
			+transferred encoder & 84.57 & 79.54 & 81.97 & 77.50&67.47& 72.13 & 75.90 & 69.00 & 72.26 & 86.05 & 79.81 & 82.79 \\
			+auxiliary learning  & 84.10 & 77.20 & 80.49 & 75.63&66.42& 70.71 & 76.31 & 68.67 & 72.29 & 86.77 & 79.46 & 82.93 \\
			LOTN & 84.00& 80.52&  82.21 &  77.08& 67.62&  72.02 &  76.61 &  70.29 &  73.29 &  86.57 &  80.89 &  83.62 \\
			\hline
	\end{tabular} }
	\centering
\end{table*}

We compare our model with the following methods:
\begin{itemize}
	\item  \textbf{Distance-rule} first performs POS tagging on the sentence, then regards the nearest adjective to the opinion target as the corresponding opinion word~\cite{DBLP:conf/kdd/HuL04}.
	\item  \textbf{Dependency-rule} collects the POS tags of opinion targets and opinion words and the dependency path between them as rule templates from the training set. Then the hign-frequency dependency templates are applied to detect the corresponding opinion words of opinion targets for the testing set~\cite{DBLP:conf/cikm/ZhuangJZ06}.
	\item  \textbf{LSTM/BiLSTM} employs word embeddings to represent words, then uses LSTM/BiLSTM network to capture the context information of input. Finally, each hidden state is fed to a softmax classifier for three-class classification to extract the opinion words of the given target~\cite{DBLP:conf/emnlp/LiuJM15}.
	\item  \textbf{Pipeline} is a combination method of BiLSTM and Distance-rule method~\cite{DBLP:conf/naacl/FanWDHC19}. It first trains a BiLSTM model on the training set. During the testing stage, the Pipeline method uses BiLSTM model to extract all the opinion words spans, then select the nearest span to the target as the extraction result.
	\item  \textbf{TC-BiLSTM} follows the design of the work for target-oriented sentiment classification~\cite{DBLP:conf/coling/TangQFL16}. This method adopts the average pooling to obtain dimension-fixed target vector, then concatenate it with word vector at each position of the sentence. Finally, the concatenation of target vectors and word vectors are fed to BiLSTM for sequence labeling.
	\item  \textbf{IOG} adopts six different positional and directional LSTMs to extract the opinion words of the target. This model achieves very powerful performance and is the state-of-the-art method in TOWE~\cite{DBLP:conf/naacl/FanWDHC19}.
	\item  \textbf{PE-BiLSTM} is our base model. It incorporates the target information into TOWE by position embedding, then uses a BiLSTM to extract the opinion words.
\end{itemize}

\subsection{Main Results}
The main experiment results are shown in Table~\ref{mainresults}. We can observe that pure rule-based methods perform very poorly compared to the supervised learning models. The method Distance-rule achieves the worst recall and F1-score in most of the datasets since it only detects the single word as opinion word.  Dependency-rule method obtains some improvements, but it is still worse than the sequence labeling models.

Compared to other neural network methods, LSTM and BiLSTM achieve poor performance because they are target-independent. They extract the same span as opinion words for different targets in a sentence. Pipeline uses BiLSTM to extract the spans of opinion words, then selects the closest span to the target as the final result. In fact, it is a target-dependent method and the distance strategy makes Pipeline method obtain nearly 20\% improvement of precision over BiLSTM in the dataset 14res. By contrast, TC-BiLSTM performs worse than Pipeline and even is inferior to BiLSTM in 14lap. One possible reason for this is that concatenating the same target vector at each position is not a good approach to incorporate the target information. IOG employs six different positional and directional LSTMs to generate rich target-dependent context representations, achieving very powerful results in all datasets. However, it also suffers from high model complexity.

PE-BiLSTM is our base method that adopts position embedding to incorporate the target information. We can observe that this simple method obtain competitive performance and is only inferior to IOG and LOTN. The experiment results show that our model LOTN achieves the best F1-score in all datasets. Compared to its base version PE-BiLSTM, LOTN obtains about 4\%$\sim$5\% improvements in F1-score. In addition, LOTN outperforms the previous state-of-the-art method IOG by 1.98\% and 2.02\% F1-score respectively in the datasets 14res and 16res. These observations demonstrate that our model can effectively transfer the latent opinions knowledge from sentiment classification datasets to the TOWE task.

\subsection{The Effects of Tranferring Encoder and Latent Opinion Words}
To investigate the effects of transferring the encoder and latent opinion words, we conduct the following experiments:

\begin{itemize}
	\item \textbf{PE-BiLSTM+transferred encoder}: We only transfer the encoder of the sentiment classification model to TOWE based on the model PE-BiLSTM.
	\item  \textbf{PE-BiLSTM+auxiliary learning}: This method only incorporates the latent opinion words from sentiment classification model into PE-BiLSTM via auxiliary learning.
\end{itemize}

Table~\ref{ablationstudy} shows the experiment results. Compared to the base model PE-BiLSTM, we can find that PE-BiLSTM+transferred encoder and PE-BiLSTM+auxiliary learning both achieve significant and consistent improvements on all datasets. The comparisons validate the effectiveness of transferring the encoder and latent opinion words from sentiment classification model for the TOWE task. After integrating both the transferred encoder and auxiliary learning, the model LOTN obtains further improvements. The results indicate that the proposed two methods are useful for the final model LOTN and they transfer opinions knowledge from different perspectives.

\subsection{The Effect of the Hyper-parameter $\lambda$}
\begin{figure}[ht]
	\centering
	\includegraphics[width=0.9\linewidth]{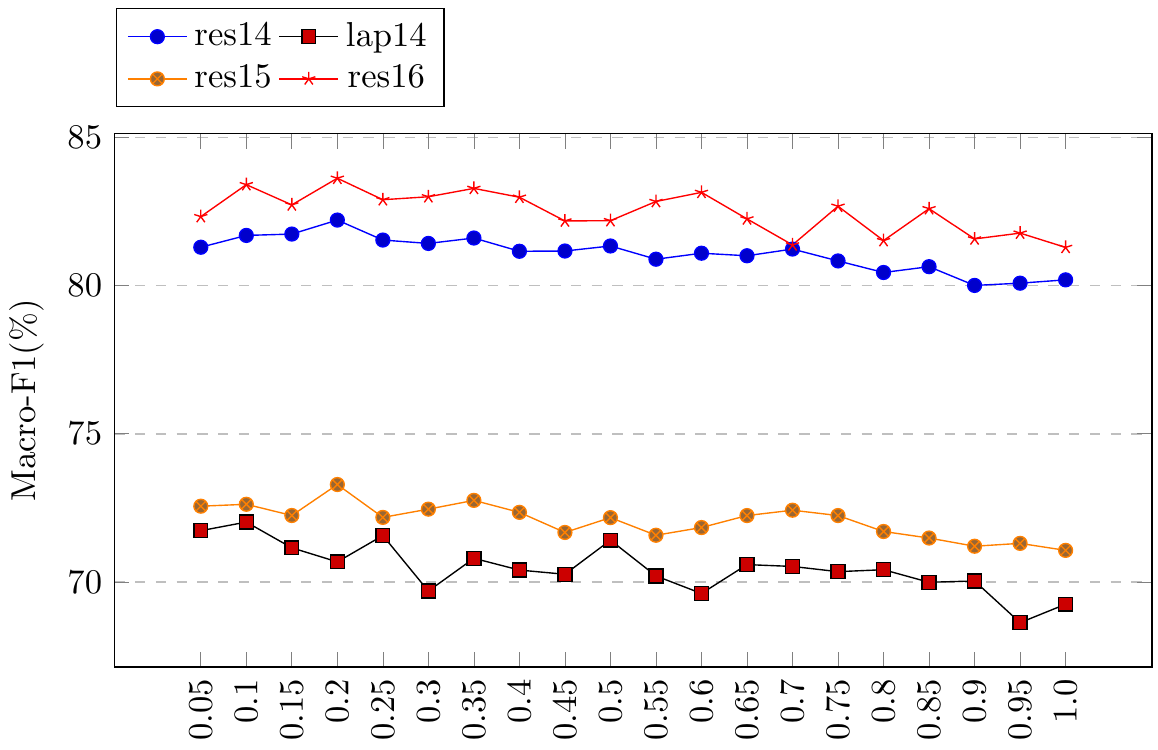}
	\caption{The effect of different hyper-parameter $\lambda$.}
	\label{hyperparameter}
\end{figure}

\noindent To analyze the effect of different $\lambda$ on our model, we adjust $\lambda$ of Equation~\ref{finalloss} in $(0, 1)$ to conduct experiments and the step is $0.05$. Figure~\ref{hyperparameter} shows the results of LOTN with different $\lambda$ on four datasets.

We can observe that LOTN achieves the relatively stable performance with varying $\lambda$ on the datasets 14res, 15res and 16res, which indicates the robustness of our method. In general, the performance of LOTN has a downward trend with an increase of $\lambda$ since the bigger $\lambda$ has a negative effect on the decoding of the model. Finally, we set $\lambda$ to be $0.1$ on 14lap and $0.2$ on other datasets.

\subsection{Case Study}
\begin{table*}[htbp]
	\centering
	\caption{Examples of the extracted results in different methods. The opinion targets are in underline and the corresponding golden opinion words are in bold.  The ``NULL'' represents that the prediction is empty.}
	\label{casestudy}
	\resizebox{0.99\textwidth}{!}
	{
		\begin{tabular}{c|c|c|c|c|cc}
			\hline 
			\multirow{2}{*}{Sentence} & \multirow{2}{*}{Distance-rule} & \multirow{2}{*}{BiLSTM} & \multirow{2}{*}{IOG} & \multirow{2}{*}{PE-BiLSTM} & \multicolumn{2}{c}{LOTN} \\ 
			\cline{6-7}
			& & & & &Latent Opinion Words & Target Decoding \\
			\hline 
			\emph{The \underline{bread} is \textbf{top notch} as well.} &  \emph{top}\XSolidBrush  & \emph{top notch}\Checkmark& \emph{top notch}\Checkmark & \emph{top notch}\Checkmark & \emph{top notch} & \emph{top notch}\Checkmark\\ 
			\hline
			
			\emph{\textbf{Great} \underline{food} but the service was dreadful!} & \emph{Great}\Checkmark & \emph{dreadful}\XSolidBrush & \emph{Great}\Checkmark &\emph{Great}\Checkmark & \emph{Great}\ & \emph{Great}\Checkmark\\ 
			\hline		
			
			\emph{Great food but the \underline{service} was \textbf{dreadful}!} & \emph{dreadful}\Checkmark & \emph{dreadful}\Checkmark & \emph{dreadful}\Checkmark &\emph{dreadful}\Checkmark & \emph{dreadful}\ & \emph{dreadful}\Checkmark\\ 
			\hline		
			
			\emph{\textbf{Good} for a \textbf{quick} \underline{sushi} lunch.} & \emph{quick}\XSolidBrush & \emph{Good, quick}\Checkmark  &  \emph{quick}\XSolidBrush  & \emph{quick}\XSolidBrush & \emph{Good} & \emph{Good, quick}\Checkmark\\ 
			\hline
			
			\emph{Their twist on pizza is healthy, but \textbf{full} of \underline{flavor}.} & \emph{full}\Checkmark & \emph{healthy}\XSolidBrush &  \emph{healthy, full}\XSolidBrush  & NULL\XSolidBrush & \emph{full} & \emph{full}\Checkmark\\ 
			\hline
		\end{tabular}
	}
\end{table*}
In order to compare different methods and validate the effectiveness of our model, we present some extracted results of the dataset 14res in Table~\ref{casestudy}.

The first example shows that the Distance-rule method cannot extract opinion phrase for the given target and thus makes the wrong prediction. Comparing the second and third examples, we can find that BiLSTM gives the same predictions for the different opinion targets since it ignores the target information. In the last two examples, the state-of-the-art model IOG and our base model PE-BiLSTM both make incorrect predictions in complicated cases, while our proposed model LOTN extracts target-dependent opinion words successfully. The results demonstrate that our model can leverage latent opinions information from sentiment classification datasets to improve the TOWE task.

\subsection{Error Analysis}
We count the distribution of different error types in the dataset 14res to analyze the weakness of LOTN. The results are given in Table~\ref{tab:erroranalysis}. The ``NULL'' represents the prediction is empty. The ``Under-extracted'' and ``Over-extracted'' respectively mean that LOTN extracts the part of the ground truth and extra words besides the golden opinion words.
\begin{table}[!htp]
	\centering
	\caption{Statistics of different error types for PE-BiLSTM and LOTN in the dataset 14res.}
	\label{tab:erroranalysis}
	\resizebox{0.98\linewidth}{!}{
		\begin{tabular}{c|c|c|c|c|c}
			\hline
			Models & {NULL} & {Under-extracted} & {Over-extracted} & {Others} & Total\\
			\hline
			PE-BiLSTM  & 76 & 107 & 49 & 34&266\\
			LOTN  & 65 & 85 & 62 & 31&243\\
			\hline
		\end{tabular}
	}
\end{table}

It can be observed that PE-BiLSTM and LOTN do not extract any opinion words in more than a quarter of error cases. In two models, the under-extracted error is the main error type. Compared to PE-BiLSTM, LOTN makes fewer mistakes in the NULL and under-extracted type. In contrast, PE-BiLSTM makes fewer over-extracted predictions. The three comparisons consistently indicate that LOTN tends to decode more opinion words under the influence of latent opinions from the sentiment classification model.

In addition, we find that the sentence is often quite long and the golden opinion words are far away from the target in NULL error. As for under-extracted cases, the models usually ignore modifiers such as the word ``\emph{most}'' in ``\emph{most delicious}'' or negators, e.g., only extracting ``\emph{best}'' from ``\emph{not the best}''. The over-extracted predictions are common in the samples containing multiple targets.

Although our model improves the overall performance of TOWE, LOTN makes some wrong predictions due to transferring some noise opinions. For example, about $5.7\%$ samples of the dataset 14res are predicted successfully by the model PE-BiLSTM but incorrectly by LOTN. The heuristic transformation method is unable to consistently yield correct opinion words for a given target, thus we inevitably introduce some noise during converting global attention knowledge into latent target-dependent opinion words.

\section{Related Work}
The early works devote to the research of opinion targets extraction, including unsupervised methods~\cite{DBLP:conf/naacl/PopescuE05,DBLP:conf/ijcnlp/0003W08,DBLP:journals/coling/QiuLBC11,DBLP:conf/emnlp/LiuXZ12} and supervised methods~\cite{jin2009novel,DBLP:conf/coling/LiHHZXZY10,DBLP:conf/emnlp/LiuJM15,DBLP:journals/kbs/PoriaCG16,DBLP:conf/acl/XuLSY18}. Recently, some works extract opinion targets and opinion words jointly in a unified framework and achieve promising results~\cite{DBLP:journals/coling/QiuLBC11,DBLP:conf/emnlp/LiuXZ12,DBLP:conf/ijcai/LiuXLZ13,DBLP:conf/emnlp/WangPDX16,DBLP:conf/aaai/WangPDX17,DBLP:conf/emnlp/LiL17}. However, these works does not extract the corresponding relation between targets and opinion words.

In fact, there are only a few works focusing on the paired opinion relations.~\citeauthor{DBLP:conf/kdd/HuL04}~\shortcite{DBLP:conf/kdd/HuL04} propose to use association rule mining for extracting opinion targets and regard the nearest adjective of targets as the corresponding opinion words.~\citeauthor{DBLP:conf/cikm/ZhuangJZ06}~\shortcite{DBLP:conf/cikm/ZhuangJZ06} adopt WordNet~\cite{miller1995wordnet} and human-built word lists to find targets and opinion words, then apply dependency-tree templates to extract the valid target-opinion pairs. These two unsupervised methods heavily depend on pre-defined rules and external resources such as syntax parser. Because of the significance of the paired relations of targets and opinion words for downstream sentiment task, ~\citeauthor{DBLP:conf/naacl/FanWDHC19}~\shortcite{DBLP:conf/naacl/FanWDHC19} propose a new subtask of ABSA, target-oriented opinion words extraction (TOWE), to extract the corresponding opinion words for a given opinion target from a review. They also design a target-fused neural sequence labeling model and achieve competitive results.

\section{Conclusion}
Insufficiency of labeled data heavily restricts the effectiveness of the neural models for TOWE. In this paper, we propose a novel model to transfer latent opinions knowledge from resource-rich review sentiment classification datasets to improve the low-resource task TOWE. Specifically, we propose an effective method to convert the attention knowledge in the sentiment classification model into target-dependent opinion words, then integrate them into TOWE via auxiliary learning signal. In addition, we also integrate the encoder of the sentiment classification model to further improve TOWE. Results from numerous experiments indicate that our approach achieves better performance than other state-of-the-art methods. Extensive analysis also demonstrates the effectiveness of our model.

\section{Acknowledgments}
We would like to thank Robert Ridley for his comments and suggestions on this paper, and the anonymous reviewers for their valuable feedback. This work was supported by the NSFC (No. 61976114, 61936012) and National Key R\&D Program of China (No. 2018YFB1005102).

\bibliography{AAAI-WuZ.7451}
\bibliographystyle{aaai}
\end{document}